\documentclass[technote,a4paper,leqno]{IEEEtran}
\pdfoutput=1
\usepackage{amssymb, amsmath}

\usepackage{filecontents}

\RequirePackage{ifpdf}
\ifpdf \PassOptionsToPackage{pdfpagelabels}{hyperref} \fi
\RequirePackage{hyperref}
\usepackage{parskip}
\usepackage[pdftex,final]{graphicx}
\usepackage{csquotes}
\usepackage{tikz}
\usepackage{tikzscale}
\usetikzlibrary{shapes, calc, shapes, arrows}
\usepackage{bm}
\usepackage{blindtext}\usepackage{braket}
\usepackage{booktabs}
\usepackage{multirow}
\usepackage{pgfplots}
\pgfplotsset{compat=newest}
\usepackage{wasysym}
\usepackage[symbol]{footmisc}
\usepackage{float}
\usepackage{caption}
\usepackage[hang]{subfigure}
\makeatletter
\newcommand\mynobreakpar{\par\nobreak\@afterheading}
\makeatother
\usepackage[noadjust]{cite}

\usepackage[nameinlink, noabbrev, capitalise]{cleveref} 
\usepackage[binary-units,group-separator={,}]{siunitx}
\DeclareSIUnit\pixel{px}
\usepackage{glossaries}
\newacronym{CNN}{CNN}{Convolutional Neural Network}
\newacronym{CUDA}{CUDA}{Compute Unified Device Architecture}
\newacronym{FCN}{FCN}{Fully Convolutional Network}
\newacronym{GPU}{GPU}{graphics processing unit}
\newacronym{LSTM}{LSTM}{Long short-term memory}
\newacronym{MLP}{MLP}{multilayer Perceptron}
\newacronym{MRF}{MRF}{Markov Random Field}
\newacronym{NLP}{NLP}{Natural Language Processing}
\newacronym{PCA}{PCA}{principal component analysis}
\newacronym{PyPI}{PyPI}{Python Package Index}
\newacronym{RNN}{RNN}{Recurrent Neural Network}
\newacronym{sst}{sst}{street segmentation toolkit}
\newacronym{GRU}{GRU}{Gated Recurrent Units}
\newacronym{DFT}{DFT}{Discrete Fourier Transformation}

\makeglossaries

\title{Creativity in Machine Learning}
\author{%
    \IEEEauthorblockN{Martin Thoma}\\
    \IEEEauthorblockA{E-Mail: info@martin-thoma.de} 
}

\hypersetup{
    pdfauthor   = {Martin Thoma},
    pdfkeywords = {recognition, machine learning, neural networks, classification, multilayer perceptron, deep learning},
    pdfsubject  = {Recognition},
    pdftitle    = {Art in Machine Learning},
}

\crefname{table}{Table}{Tables}
\crefname{figure}{Figure}{Figures}

\usepackage{microtype}

\begin{document}
\maketitle
\begin{abstract}
Recent machine learning techniques can be modified to produce creative
results. Those results did not exist before; it is not a trivial
combination of the data which was fed into the machine learning system. The
obtained results come in multiple forms: As images, as text and as audio.

This paper gives a high level overview of how they are created and gives some
examples. It is meant to be a summary of the current work and give people
who are new to machine learning some starting points.
\end{abstract}


\section{Introduction}
\label{sec:introduction}
According to \cite{LongmanDCE06} \textit{creativity} is \enquote{the ability to
use your imagination to produce new ideas, make things etc.} and
\textit{imagination} is \enquote{the ability to form pictures or ideas in your
mind}.

Recent advances in machine learning produce results which the author would
intuitively call creative. A high-level overview over several of those
algorithms are described in the following.

This paper is structured as follows: \Cref{sec:ml-basics} introduces the
reader on a very simple and superficial level to machine learning,
\cref{sec:images} gives examples of creativity with images,
\cref{sec:text-generation} gives examples of machines producing textual
content, and \cref{sec:music} gives examples of machine learning and music. A
discussion follows in
\cref{sec:discussion}.


\section{Basics of Machine Learning}
\label{sec:ml-basics}
The traditional approach of solving problems with software is to program
machines to do so. The task is divided in as simple sub-tasks as possible,
the subtasks are analyzed and the machine is instructed to process the input
with human-designed algorithms to produce the desired output. However, for
some tasks like object recognition this approach is not feasible. There are
way to many different objects, different lighting situations, variations in
rotation and the arrangement of a scene for a human to think of all of them and
model them. But with the internet, cheap computers, cameras, crowd-sourcing
platforms like Wikipedia and lots of Websites, services like Amazon Mechanical
Turk and several other changes in the past decades a lot of data has become
available. The idea of machine learning is to make use of this data.

A formal definition of the field of Machine Learning is given by
Tom~Mitchel~\cite{Mitchell97}:
\begin{displayquote}
A computer program is said to learn from experience~$E$ with respect to some
class of tasks~$T$ and performance measure~$P$, if its performance at tasks
in~$T$, as measured by~$P$, improves with experience~$E$.
\end{displayquote}

This means that machine learning programs adjust internal parameters to fit the
data they are given. Those computer programs are still developed by software
developers, but the developer writes them in a way which makes it possible to
adjust them without having to re-program everything. Machine learning programs
should generally improve when they are fed with more data.

The field of machine learning is related to statistics. Some algorithms
directly try to find models which are based on well-known distribution
assumptions of the developer, others are more general.

A common misunderstanding of people who are not related in this field is that
the developers don't understand what their machine learning program is doing.
It is understood very well in the sense that the developer, given only a pen,
lots of paper and a calculator could calculate the same result as the machine
does when he gets the same data. And lots of time, of course. It is not
understood in the sense that it is hard to make predictions how the algorithm
behaves without actually trying it. However, this is similar to expecting from
an electrical engineer to explain how a computer works. The electrical engineer
could probably get the knowledge he needs to do so, but the amount of time
required to understand such a complex system from basic building blocks is
a time-intensive and difficult task.

\begin{figure}
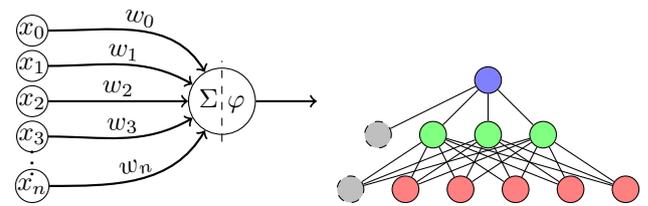

\centering
\subfigure[Example of an artificial neuron unit. $x_i$ are the input signals and $w_i$ are weights which have to get learned. Each input signal gets multiplied with its weight, everything gets summed up and the activation function $\varphi$ is applied.]{
  \label{fig:artificial-neuron}
  \includegraphics[width=0.45\linewidth]{neuron.tikz}
}%
\subfigure[A visualization of a simple feed-forward neural network. The 5~input nodes are red, the 2~bias nodes are gray, the 3~hidden units are green and the single output node is blue.]{
  \label{fig:feed-forward-nn}
  \includegraphics[width=0.45\linewidth]{feed-forward-nn.tikz}
}
\caption{Neural networks are based on simple units which get combined to complex networks.}
\label{fig:neural-style}
\end{figure}

An important group of machine learning algorithms was inspired by biological
neurons and are thus called \textit{artificial neural networks}. Those networks
are based on mathematical functions called \textit{artificial neurons} which
take $n \in \mathbb{N}$ numbers $x_1, \dots, x_n \in \mathbb{R}$ as input,
multiply them with weights $w_1, \dots, w_n \in \mathbb{R}$, add them and apply
a so called \textit{activation function} $\varphi$ as visualized in
\cref{fig:artificial-neuron}. One example of such an activation function is the
sigmoid function $\varphi(x) = \frac{1}{1+e^{-x}}$. Those functions act as
building blocks for more complex systems as they can be chained and grouped in
layers as visualized in \cref{fig:feed-forward-nn}. The interesting question is
how the parameters $w_i$ are learned. This is usually done by an optimization
technique called
\textit{gradient descent}. The gradient descent algorithm takes a function
which has to be derivable, starts at any point of the surface of this error
function and makes a step in the direction which goes downwards. Hence it tries
to find a minimum of this high-dimensional function.

There is, of course, a lot more to say about machine learning. The interested
reader might want to read the introduction given by Mitchell~\cite{Mitchell97}.


\section{Image Data}\label{sec:images}%
Applying a simple neural network on image data directly can work, but the number
of parameters gets extraordinary large. One would take one neuron per pixel and
channel. This means for $\SI{500}{\pixel} \times \SI{500}{\pixel}$ RGB images
one would get \num{750000} input signals. To approach this problem, so called
\glspl{CNN} were introduced. Instead of learning the full connection between
the input layer and the first hidden layer, those networks make use of
convolution layers. Convolution layers learn a convolution; this means they
learn the weights of an image filter. An additional advantage is that
\glspl{CNN} make use of spacial relationships of the pixels instead of
flattening the image to a stream of single numbers.

An excellent introduction into \glspl{CNN} is given by~\cite{Nielsen2015}.

\subsection{Google DeepDream}\label{subsec:google-deepdream}%
The gradient descent algorithm which optimizes most of the parameters in neural
networks is well-understood. However, the effect it has on the recognition
system is difficult to estimate. \cite{inceptionism2015} proposes a technique
to analyze the weights learned by such a network. A similar idea was applied
by~\cite{vondrick2013hoggles}.

For example, consider a neural network which was trained to recognize various
images like bananas. This technique turns the network upside down and starts
with random noise. To analyze what the network considers bananas to look like,
the random noise image is gradually tweaked so that it generates the output
\enquote{banana}. Additionally, the changes can be restricted in a way that the
statistics of the input image have to be similar to natural images. One example
of this is that neighboring pixels are correlated.
\goodbreak
Another technique is to amplify the output of layers. This was described
in~\cite{inceptionism2015}:\nobreak%
\begin{displayquote}
We ask the network: \enquote{Whatever you see there, I want more of it!} This
creates a feedback loop: if a cloud looks a little bit like a bird, the network
will make it look more like a bird. This in turn will make the network
recognize the bird even more strongly on the next pass and so forth, until a
highly detailed bird appears, seemingly out of nowhere.
\end{displayquote}

\begin{figure}[ht]
    \centering
    \includegraphics[width=0.45\textwidth]{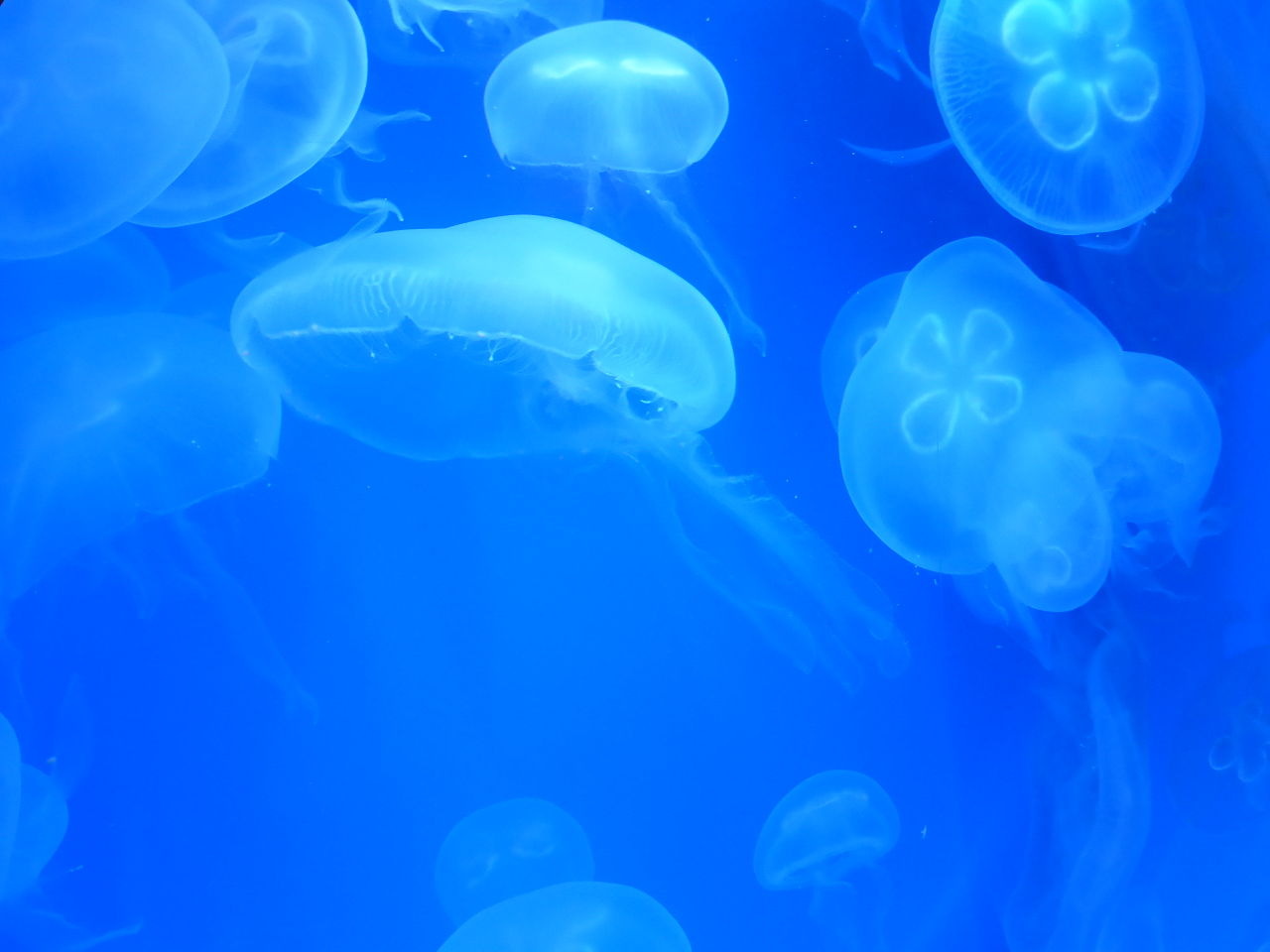}
    \caption{Aurelia aurita}
    \label{fig:Aurelia-aurita-3-original}
\end{figure}

\begin{figure}[ht]
    \centering
    \includegraphics[width=0.45\textwidth]{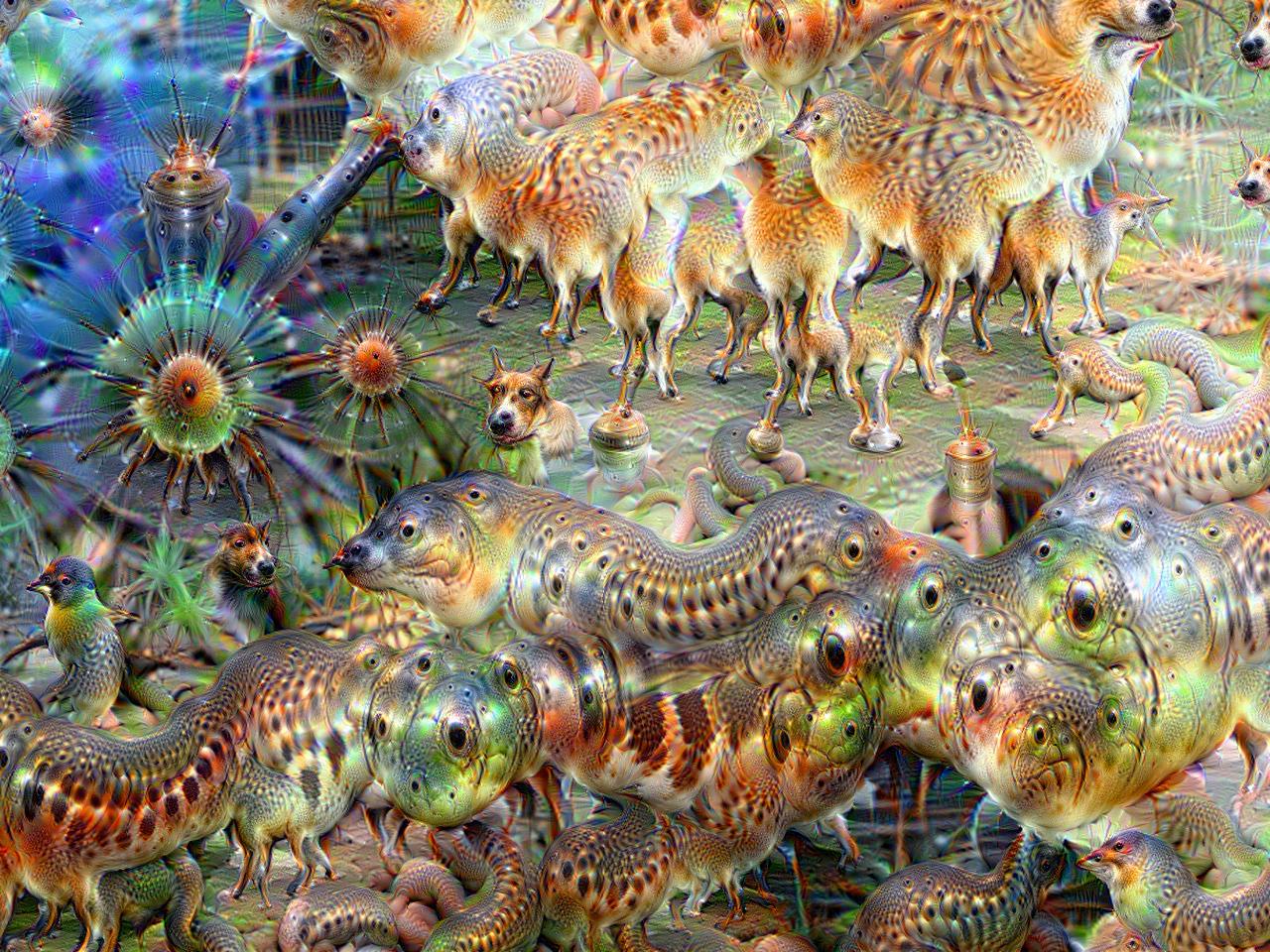}
    \caption{DeepDream impression of Aurelia aurita}
    \label{fig:Aurelia-aurita-3-100}
\end{figure}

The name \enquote{Inceptionism} in the title of~\cite{inceptionism2015} comes
from the science-fiction movie \enquote{Inception}~(2010). One reason it might
be chosen is because neural networks are structured in layers. Recent
publications tend to have more and more layers~\cite{he2015deep}. The used
jargon is to say they get \enquote{deeper}. As this technique as published by
Google engineers, the technique is called \textit{Google DeepDream}.

It has become famous in the internet~\cite{RedditDeepDream}. Usually, the
images are generated in iterations and in each iteration it is zoomed into the
image.\\
Images and videos published by the Google engineers can be seen
at~\cite{googlePhotos2015}. \Cref{fig:Aurelia-aurita-3-original} shows the
original image from which \cref{fig:Aurelia-aurita-3-100} was created with the
deep dream algorithm.


\subsection{Artistic Style Imitation}
A key idea of neural networks is that they learn different representations of
the data in each layer. In the case of \glspl{CNN}, this can easily be
visualized as it was done in various papers~\cite{zeiler2014visualizing}.
Usually, one finds that the network learned to build edge detectors in the
first layer and more complex structures in the upper layers.

\begin{figure}
\centering
\subfigure[Original Image]{
  \label{fig:scottish-highland-cattle}
  \includegraphics[width=0.45\linewidth, keepaspectratio]{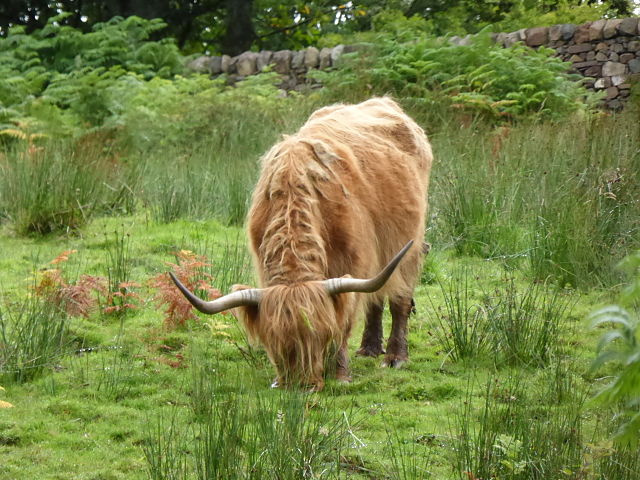}
}%
\subfigure[Style image]{
  \label{fig:starry-night}
  \includegraphics[width=0.45\linewidth, keepaspectratio]{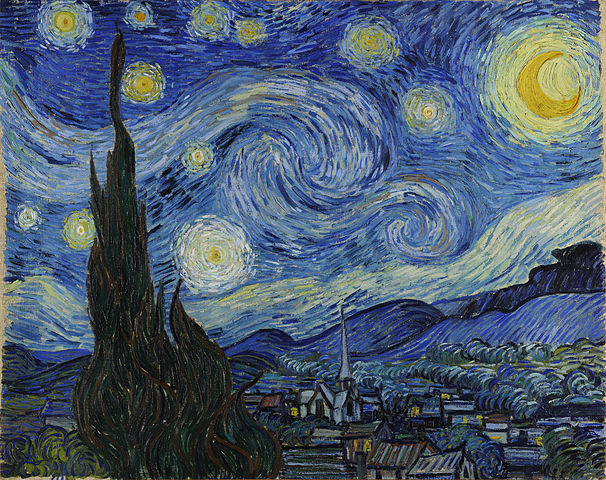}
}

\subfigure[The artistic style of Van Gogh's \enquote{Starry Night} applied to the photograph of a Scottish Highland Cattle.]{
  \label{fig:highland-van-gogh}
  \includegraphics[width=0.93\linewidth, keepaspectratio]{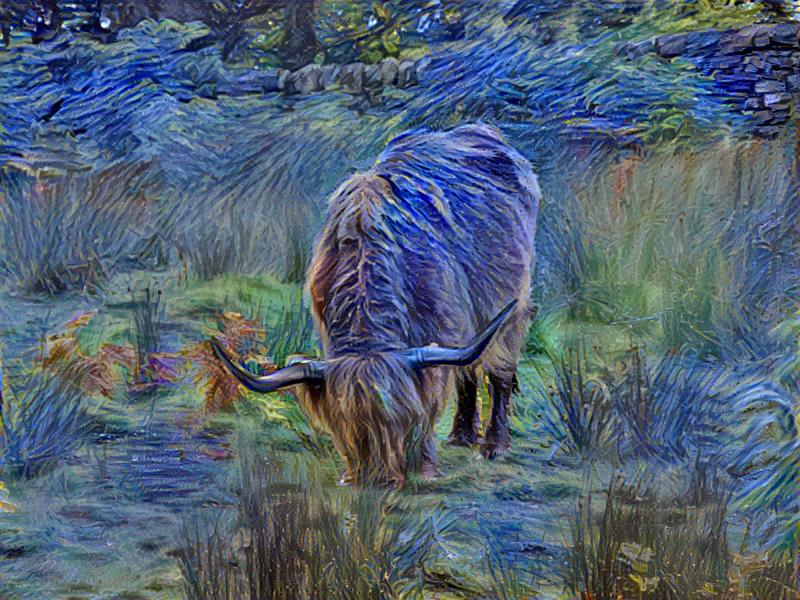}
}
\caption{The algorithm takes both, the original image and the style image  to produce the result.}
\label{fig:neural-style}
\end{figure}

Gatys, Ecker and Bethge showed in~\cite{gatys2015neural} that with a clever
choice of features it is possible to separate the general style of an image in
terms of local image appearance from the content of an image. They support
their claim by applying the style of different artists to an arbitrary image of
their choice.

This artistic style imitation can be seen itself as creative work. An example
is given by~\cref{fig:neural-style}. The code which created this example is
available under~\cite{Johnson2016}.

Something similar was done by~\cite{shih2014style}, where the style of a
portrait photograph was transferred to another photograph. A demo can be seen
on~\cite{Shih2014}.

\subsection{Drawing Robots}
Patrick Tresset and Frédéric Fol Leymarie created a system called AIKON
(Automatic IKONic drawing) which can automatically generated sketches for
portraits~\cite{tresset2005generative}. AIKON takes a digital photograph,
detects faces on them and sketches them with a pen-plotter.

Tresset and Leymaire use $k$-means clustering~\cite{1017616} to segment regions
of the photograph with similar color which, in turn, will get a similar
shading.

Such a drawing robot could apply machine learning techniques known from
computer vision for detecting the human. It could apply self-learning
techniques to draw results most similar to the artists impression of the image.
However, the system described in~\cite{tresset2005generative} seems not to be a
machine learning computer program according to the definition by Tom
Mitchell~\cite{Mitchell97}.


\section{Text Data}%
\label{sec:text-generation}%

Digital text is the first form of natural communication which involved
computers. It is used in the form of chats, websites, on collaborative projects
like Wikipedia, in scientific literature. Of course, it was used in pre-digital
times, too: In newspaper, in novels, in dramas, in religious texts like the
bible, in books for education, in notes from conversations.

This list could be continued and most of these kinds of texts are now available
in digital form. This digital form can be used to teach machines to generate
similar texts.

The most simple language model which is of use is an $n$-gram model. This model
makes use of sequences of the length $n$ to model language. It can be used to
get the probability of a third word, given the previous two words. This way, a
complete text can be generated word by word. Refinements and extensions to this
model are discussed in the field of \gls{NLP}.

However, there are much more sophisticated models. One of those are
\textit{character predictors} based on \glspl{RNN}. Those character predictors
take a sequence of characters as input and predict the next character. In that
sense they are similar to the $n$-gram model, but operate on a lower level.
Using such a predictor, one can generate texts character by character. If the
model is good, the text can have the correct punctuation. This would not be
possible with a word predictor.

Character predictors can be implemented with \glspl{RNN}. In contrast to
standard feed-forward neural networks like \glspl{MLP} which was shown in
\cref{fig:feed-forward-nn}, those networks are trained to take their output at
some point as well as the normal input. This means they can keep some
information over time. One of the most common variant to implement \glspl{RNN}
is by using so called \gls{LSTM} cells~\cite{hochreiter1997long}.

Recurrent networks apply two main ideas in order to learn: The first is called
\textit{unrolling} and means that an recurrent network is imagined to be
an infinite network over time. At each time step the recurrent neurons get
duplicated. The second idea is \textit{weight sharing} which means that those
unrolled neurons share the same weight.

\subsection{Similar Texts Generation}
Karpathy trained multiple character \glspl{RNN} on different datasets and gave
an excellent introduction~\cite{Karpathy2015}. He trained it on Paul Graham's
essays, all the works of Shakespeare, the Hutter Prize~\cite{hutterPrize}
\SI{100}{\mega\byte}~dataset of raw Wikipedia articles, the raw \LaTeX{} source
file of a book about algebraic stacks and geometry and Linux C~code.

With that training data, the models can generate similar texts. New works which
look like Shakespeare plays, new Wikipedia articles, new Linux code and new
papers about algebraic geometry can thus automatically be generated. At a first
glance, they do look authentic. The syntax was mostly used correctly, the
formatting looks as expected, the sentences are grammatically correct. However,
when one looks at the broader context it is easy to recognize that the
algorithm has no insight in what it is doing. It does match patterns really
well, but it fails to follow a central theme. In the context of C~code this
means that new variables are introduced, but not used. At the same time,
variables which were not declared are used. In the context of Shakespear plays
this means that a lot of new characters are introduced, but they don't speak
with each other or about each other.

The code used to generate these examples is available and ready to use
through~\cite{charRNNgithub}. A couple of examples are
in~\cref{sec:appendix-auto-generated-text}.

\subsection{Chatbots}%
\label{subsec:chatbots}%

Chatbots are computer programs which participate in chat rooms as autonomous
agents. This means they have similar permissions and possibilities as usual human
users have, but users can trigger a set of commands to make the bot give them
valuable information or features.

A special category of chatbots are such bots which actively participate in the
conversation, which is usually not the case. One of the earliest programs in
this category is ELIZA, a bot created by Joseph Weizenbaum in the
1960s~\cite{Weizenbaum1976}. This program had a set of patterns implemented to
which it would reply in a seemingly smart way in a psychologists fashion. This
means quite often the program would simply repeat the last sentence and add
something meaningless like \enquote{How do you feel about it?}. According
to~\cite{Curtis2014}, Weizenbaum once found his secretary --- who was aware of
the fact that this is a computer program --- chatting with the machine. When he
looked over her shoulder, she asked him \enquote{would you mind leaving the
room}.

Today, much more sophisticated chatbots exist. They make use of the vast amount
of data which is available by millions of Wikipedia articles, chat protocols,
websites, help desk protocols, subtitles of movies as well as the astonishing
increase in computing power to train
\glspl{RNN} and language models similar to the ones described before.

Interesting results like the following were obtained
by~\cite{vinyals2015neural}:
\begin{displayquote}
\textbf{Human}: what is the purpose of life ?\\
\textbf{Machine}: to serve the greater good .\\
\textbf{Human}: what is the purpose of living ?\\
\textbf{Machine}: to live forever .
\end{displayquote}


\section{Audio Data}
\label{sec:music}

Common machine learning tasks which involve audio data are speech recognition,
speaker identification, identification of songs. This leads to some
less-common, but interesting topics: The composition of music, the synthesizing
of audio as art. While the composition might be considered in
\cref{sec:text-generation}, we will now investigate the work which was done in
audio synthesization.

\subsection{Emily Howell}
David Cope created a project called \enquote{Experiments in Musical
Intelligence} (short: EMI or Emmy) in 1984~\cite{Cope1987}. He introduces the idea of
seeing music as a language which can be analyzed with natural language
processing (NLP) methods. Cope mentions that EMI was more useful to him, when
he used the system to \enquote{create small phrase-size textures as next
possibilities using its syntactic dictionary and rule base}~\cite{Cope1987}.

In 2003, Cope started a new project which was based on EMI: Emily
Howell~\cite{cope2013well}. This program is able to \enquote{creat[e] both
highly authentic replications and novel music compositions}. The reader might
want to listen to~\cite{Cope2012} to get an impression of the beauty of the
created music.

According to Cope, an essential part of music is \enquote{a set of instructions
for creating different, but highly related self-replications}. Emmy was
programmed to find this set of instructions. It tries to find the
\enquote{signature} of a composer, which Cope describes as \enquote{contiguous
patterns that recur in two or more works of the composer}.

The new feature of \textit{Emily Howell} compared to \textit{Emmy} is that
Emily Howell does not necessarily remain in a single, already known style.

Emily Howell makes use of association network. Cope emphasizes that this is not
a form of a neural network. However, it is not clear from~\cite{cope2013well}
how exactly an association network is trained. Cope mentions that Emily
Howell is explained in detail in~\cite{cope2005computer}.

\subsection{GRUV}

Recurrent neural networks --- \gls{LSTM} networks, to be exact --- are used
in~\cite{nayebigruv} together with \gls{GRU} to build a network which can be
trained to generate music. Instead of taking notes directly or MIDI files,
Nayebi and Vitelli took raw audio waveforms as input. Those audio waveforms are
feature vectors given for time steps $0, 1, \dots, t-1, t$. The network is
given those feature vectors $X_1, \dots, X_t$ and has to predict the following
feature vector $X_{t+1}$. This means it continues the music. As the input is
continuous, the problem was modeled as a regression task. \Gls{DFT} was used on
chunks of length $N$ of the music to obtain features in the frequency domain.

An implementation can be found at~\cite{gruvGitHub} and a demonstration can
be found at~\cite{Vitelli2015}.

\subsection{Audio Synthesization}
Audio synthesization is generating new audio files. This can either be music or
speech. With the techniques described before, neural networks can be trained to
generate music note by note. However, it is desirable to allow multiple notes
being played at the same time.

This idea and some others were applied by Daniel Johnson. He wrote a very good
introduction into neural networks for music composition which explains those
ideas~\cite{Johnson2015}. Example compositions are available there, too. He
also made the code for his Biaxial Recurrent Neural Network available
under~\cite{Johnson2015a}.






\section{Discussion}
\label{sec:discussion}

What does these examples mean for our understanding of creativity? Does it
influence how much we value art? Could we define art and creativity better
after having those and similar results?

I think we might readjust our understanding of creativity just like we adjusted
our understanding of algorithmically hard problems after Deep Blue won against
the reigning world chess champion Garry Kasparov in~1997.

However, by now it is obvious that machine learning algorithms cannot compete
with human artists. Today's state of the art algorithms which are purely based
on machine learning don't follow a central theme. They lack the ability to
plan. Although clever algorithms were implemented for composing music, it seems
as if there is still a lot of supervision involved.

\bibliographystyle{IEEEtranSA}
\bibliography{art-in-machine-learning}
\printglossary


\clearpage\onecolumn
\begin{appendices}
\section{Automatically generated Texts}\label{sec:appendix-auto-generated-text}
The following texts were generated by~\cite{charRNNgithub} and published by
Karpathy on~\cite{Karpathy2015}. This is meant to be a copy for convenience
and the case that the website is not reachable.

\subsection{Shakespeare}
\begin{verbatim}
PANDARUS:
Alas, I think he shall be come approached and the day
When little srain would be attain'd into being never fed,
And who is but a chain and subjects of his death,
I should not sleep.

Second Senator:
They are away this miseries, produced upon my soul,
Breaking and strongly should be buried, when I perish
The earth and thoughts of many states.

DUKE VINCENTIO:
Well, your wit is in the care of side and that.

Second Lord:
They would be ruled after this chamber, and
my fair nues begun out of the fact, to be conveyed,
Whose noble souls I'll have the heart of the wars.

Clown:
Come, sir, I will make did behold your worship.

VIOLA:
I'll drink it.
\end{verbatim}

\end{appendices}

\subsection{Wikipeda}
\begin{verbatim}
Naturalism and decision for the majority of Arab countries' capitalide was grounded
by the Irish language by [[John Clair]], [[An Imperial Japanese Revolt]], associated
with Guangzham's sovereignty. His generals were the powerful ruler of the Portugal
in the [[Protestant Immineners]], which could be said to be directly in Cantonese
Communication, which followed a ceremony and set inspired prison, training. The
emperor travelled back to [[Antioch, Perth, October 25|21]] to note, the Kingdom
of Costa Rica, unsuccessful fashioned the [[Thrales]], [[Cynth's Dajoard]], known
in western [[Scotland]], near Italy to the conquest of India with the conflict.
Copyright was the succession of independence in the slop of Syrian influence that
was a famous German movement based on a more popular servicious, non-doctrinal
and sexual power post. Many governments recognize the military housing of the
[[Civil Liberalization and Infantry Resolution 265 National Party in Hungary]],
that is sympathetic to be to the [[Punjab Resolution]]
(PJS)[http://www.humah.yahoo.com/guardian.
cfm/7754800786d17551963s89.htm Official economics Adjoint for the Nazism, Montgomery
was swear to advance to the resources for those Socialism's rule,
was starting to signing a major tripad of aid exile.]]
\end{verbatim}

\subsection{Linux Code, 1}
\begin{verbatim}
/*
 * Increment the size file of the new incorrect UI_FILTER group information
 * of the size generatively.
 */
static int indicate_policy(void)
{
  int error;
  if (fd == MARN_EPT) {
    /*
     * The kernel blank will coeld it to userspace.
     */
    if (ss->segment < mem_total)
      unblock_graph_and_set_blocked();
    else
      ret = 1;
    goto bail;
  }
  segaddr = in_SB(in.addr);
  selector = seg / 16;
  setup_works = true;
  for (i = 0; i < blocks; i++) {
    seq = buf[i++];
    bpf = bd->bd.next + i * search;
    if (fd) {
      current = blocked;
    }
  }
  rw->name = "Getjbbregs";
  bprm_self_clearl(&iv->version);
  regs->new = blocks[(BPF_STATS << info->historidac)] | PFMR_CLOBATHINC_SECONDS << 12;
  return segtable;
}
\end{verbatim}

\subsection{Linux Code, 2}
\begin{verbatim}
/*
 *  Copyright (c) 2006-2010, Intel Mobile Communications.  All rights reserved.
 *
 *   This program is free software; you can redistribute it and/or modify it
 * under the terms of the GNU General Public License version 2 as published by
 * the Free Software Foundation.
 *
 *        This program is distributed in the hope that it will be useful,
 * but WITHOUT ANY WARRANTY; without even the implied warranty of
 *   MERCHANTABILITY or FITNESS FOR A PARTICULAR PURPOSE.  See the
 *
 *  GNU General Public License for more details.
 *
 *   You should have received a copy of the GNU General Public License
 *    along with this program; if not, write to the Free Software Foundation,
 *  Inc., 675 Mass Ave, Cambridge, MA 02139, USA.
 */

#include <linux/kexec.h>
#include <linux/errno.h>
#include <linux/io.h>
#include <linux/platform_device.h>
#include <linux/multi.h>
#include <linux/ckevent.h>

#include <asm/io.h>
#include <asm/prom.h>
#include <asm/e820.h>
#include <asm/system_info.h>
#include <asm/setew.h>
#include <asm/pgproto.h>

#define REG_PG    vesa_slot_addr_pack
#define PFM_NOCOMP  AFSR(0, load)
#define STACK_DDR(type)     (func)

#define SWAP_ALLOCATE(nr)     (e)
#define emulate_sigs()  arch_get_unaligned_child()
#define access_rw(TST)  asm volatile("movd %%esp, %0, %3" : : "r" (0));   \
  if (__type & DO_READ)

static void stat_PC_SEC __read_mostly offsetof(struct seq_argsqueue, \
          pC>[1]);

static void
os_prefix(unsigned long sys)
{
#ifdef CONFIG_PREEMPT
  PUT_PARAM_RAID(2, sel) = get_state_state();
  set_pid_sum((unsigned long)state, current_state_str(),
           (unsigned long)-1->lr_full; low;
}
\end{verbatim}

\end{document}